\documentclass[letterpaper, 10 pt, conference]{ieeeconf}  

\IEEEoverridecommandlockouts                              

\usepackage{graphicx} 
\usepackage{url} 
\title{\LARGE \bf
Autonomous Driving using Spiking Neural Networks on Dynamic Vision Sensor Data: A Case Study of Traffic Light Change Detection
}

\author{Xuelei Chen$^{1}$ and Sotirios Spanogianopoulos$^{2}$
\thanks{$^{1}$Xuelei Chen with Boston University, Boston, MA 02215, USA
        {\tt\small xuelei@bu.edu}}%
\thanks{$^{2}$ Sotirios Spanogianopoulos with University of Portsmouth, Portsmouth, UK
        {\tt\small sotiris.spanogianopoulos@gmail.com}}%
}

\begin{document}

\maketitle
\thispagestyle{empty}
\pagestyle{empty}

\begin{abstract}

Autonomous driving is a challenging task that has gained broad attention from both academia and industry. Current solutions using convolutional neural networks require large amounts of computational resources, leading to high power consumption. Spiking neural networks (SNNs) provide an alternative computational model to process information and make decisions. This biologically plausible model has the advantage of low latency and energy efficiency. Recent work using SNNs for autonomous driving mostly focused on simple tasks like lane keeping in simplified simulation environments. This paper studies SNNs on photo-realistic driving scenes in the CARLA simulator, which is an important step toward using SNNs on real vehicles. The efficacy and generalizability of the method will be investigated.

\end{abstract}

\section{INTRODUCTION}\label{sec:intro}

Autonomous driving is a promising technology that can make future road transportation more convenient and accessible. An autonomous driving system usually contains multiple components like perception, prediction, planning, and control modules. Deep neural networks are a powerful computation model that plays important roles in all these modules. However, most current methods rely on artificial neural networks or convolutional neural networks, which demand a lot of computational resources. This can cause latency and high energy consumption.

Biologically inspired spiking neural networks (SNNs) \cite{gerstner2002spiking} can circumvent these problems. SNNs are sometimes also referred to as the third generation of the neural network. The neurons in SNNs mimic the mechanism of neurons in human brains. The input data of SNNs are spike sequences compared to static input data of ANN or CNN. This brings the benefit of leveraging temporal information. Another benefit of energy efficiency comes from the sparse coding and event-driven processing in SNNs. Some companies have released neuromorphic chips, like IBM TrueNorth and Intel Loihi, which are specially designed for SNNs.

Dynamic vision sensors (DVSs) or event cameras are a new kind of sensors that detect only changes in the scene and were first invented in 2008 \cite{lichtsteiner2008128}. DVS data can be good input data for SNNs without the need for complicated preprocessing of data conversion, which is usually required if using SNNs on RGB image data.

There has been some research on SNNs in the robotics community because of SNNs' low-power hardware implementation with technologies like memristor, and advantages like explainability and biological plausibility. Recent works combining SNNs and DVS data include \cite{massa2020efficient,gehrig2020event,cordone2022object}, which solved the problems of gesture recognition, road object detection, and UAV angular velocity regression. We follow this trend to investigate further end-to-end decision-making using SNNs and DVS data.

The contributions of this paper are: 
(1) To the best of our knowledge, this is the first research that investigates SNNs with DVS data for photo-realistic autonomous driving in CARLA. 
(2) The generalizability of SNNs in urban driving scenes will be investigated. 
(3) An SNN-based traffic light change detection method is implemented.

\section{RELATED WORK}

\subsection{SNNs for Perception and Control}
SNNs have proven to be effective in both robotic perception \cite{hussaini2022spiking} and sensorimotor control \cite{bing2018survey}. Autonomous driving, as a sub-field of robotics research, has also experienced a surge in the popularity of SNNs \cite{DBLP:conf/visapp/MohapatraGYMZ19}. An object detection framework using SNNs on LiDAR data has been proposed in \cite{wang2021temporal}. In \cite{meschede2017training} and\cite{bing2018end}, the SNN was trained for a lane-keeping vehicle using DVS data. However, their simulation environment only has the vehicle itself and white lane markings on the gray ground, which is much simpler than real-life autonomous driving. Loeve \cite{loeve2020optimizing} implemented SNNs for autonomous driving in photo-realistic simulator CARLA \cite{dosovitskiy2017carla}. However, the implemented SNNs required manual tuning of the parameters, and the input data were segmented camera data. The method's usability on real autonomous vehicles is restricted by these limitations.

\subsection{Software for SNNs}
Early papers mostly wrote code based on their own tasks and models. It takes a long time to adapt their code for other tasks.

There are also several open-sourced software for SNN building and training since 2014. Nengo \cite{bekolay2014nengo} is a pioneer work that provides Python tools not only for SNNs but also for many other brain modeling methods. SpikingJelly \cite{SpikingJelly} is a recently published library based on PyTorch and can be used in almost the same way as the original PyTorch. snnTorch \cite{eshraghian2021training} is another recently published library, which is also based on PyTorch and its usage is very similar to SpikingJelly. CARLsim 6.0 \cite{niedermeier2022carlsim} is a GPU-accelerated SNN simulator that has a C/C++ interface.
This project will use SpikingJelly to build and train the SNN.

\subsection{DVS Dataset}
The dynamic vision sensor (DVS) is a young product in the market, and researchers from the fields of robotics and computer vision started working on DVS data just a few years ago. A series of works that use DVS data generated and processed from conventional images will not be listed here. Some real DVS datasets captured by DVSs include: DVS128 Gesture \cite{amir2017low}, ASL-DVS \cite{bi2019graph}, DSEC \cite{gehrig2021dsec} and GEN1 \cite{de2020large}. DSEC and GEN1 are collected on real roads aiming to solve different perception tasks in autonomous driving.

CARLA simulator also provides dynamic vision sensors that can be mounted on the simulated vehicles. This project will use data collected from CARLA to solve an SNN-based end-to-end vehicle decision-making according to traffic lights in the crosswalk.

\section{PRELIMINARIES}

\subsection{SNN Models}
The key component in SNNs is the spiking neuron. Each spiking neuron fires when its membrane potential reaches a threshold. The emitted signal will increase or decrease the potential of its connected postsynaptic neuron. The most widely used neuron model in SNNs is described by the Leaky-Integrate-and-Fire (LIF) dynamics, which is formulated in the following equation:
\begin{equation}
    \tau\frac{dV}{dt} = (E_r - V) + \frac{I}{g_L}
\end{equation}
where $V$ is the membrane potential, $\tau$ is the time constant of the membrane, $E_r$ is the resting potential, $I$ is the total external current from all parent presynaptic neurons, and $g_L$ is the leaky conductance.
The input of the SNN and the output of all intermediate neurons are spike sequences which can be described as 
\begin{equation}
    S(t) = \sum_{i}\delta(t-t_i)
\end{equation}
where $t_i$ is the time when the $i$-th spike occurs. 

The current incurred by the presynaptic neuron can be described using the exponential function
\begin{equation}
I(t) = \int_0^{\infty}S(s-t)\exp(-s/\tau_s)ds
\end{equation}
where $\tau_s$ is the synaptic time constant.
\subsection{SNN Training}
Many training methods have been proposed for SNN parameter optimization. The earliest method is Hebbian learning \cite{hebb2005organization}. Spike-Timing-Dependent-Plasticity (STDP) is a modified version of Hebbian learning and uses the temporal information of presynaptic and postsynaptic spikes to update the synaptic weight. More specifically, if a postsynaptic spike occurs immediately after a presynaptic spike, then the weight between these two neurons will be increased. Otherwise, the weight will be decreased. These training methods do not require gradient calculation, thus not affected by the non-differentiability of spiking neurons.

Due to the success of deep learning in many fields, we also want to use gradient-based optimization techniques like back-propagation on SNNs. A workaround is to use a differentiable function to approximate the binary all-or-nothing-style output of SNNs. Some methods include spike response models (SRM) and adaptive exponential integrate-and-fire (AdEx) models. Another way is to use a surrogate gradient instead of changing the whole model. A detailed review of commonly used surrogate gradients in SNNs can be found in \cite{neftci2019surrogate}.

\subsection{DVS Data}
\begin{figure}[ht]
    \centering
    \includegraphics[width=0.9\linewidth]{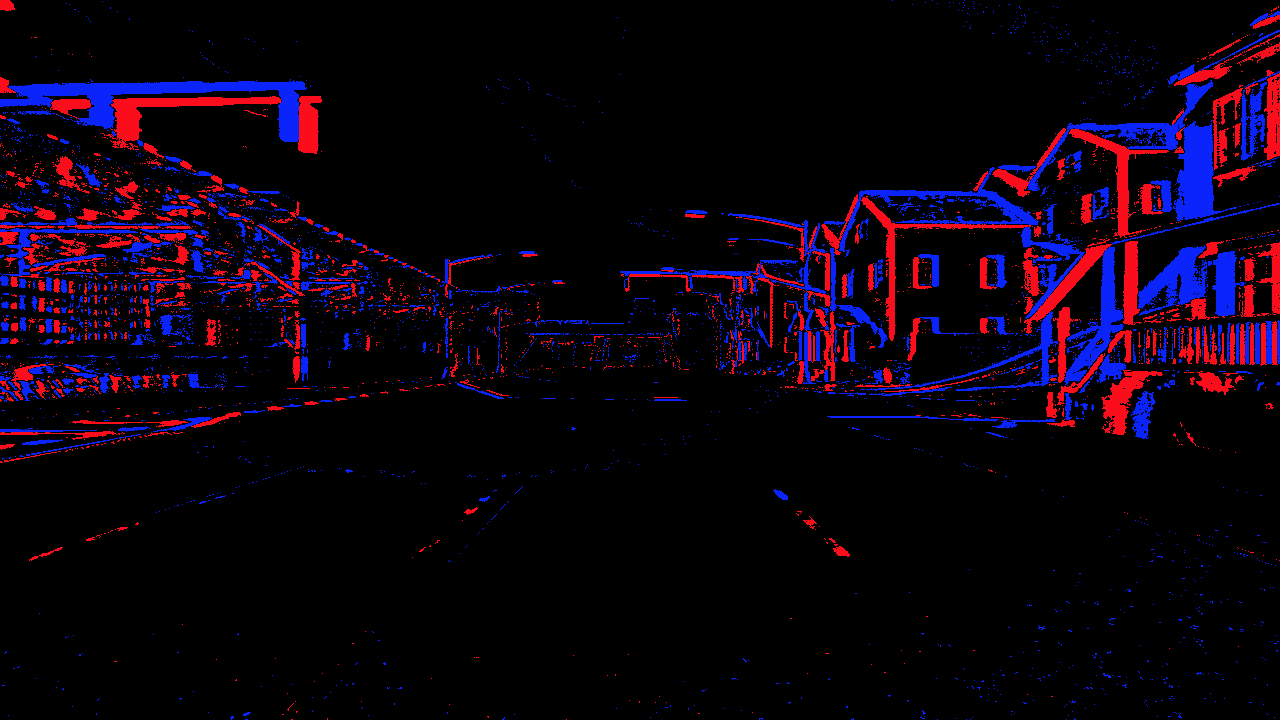}
    \includegraphics[width=0.9\linewidth]{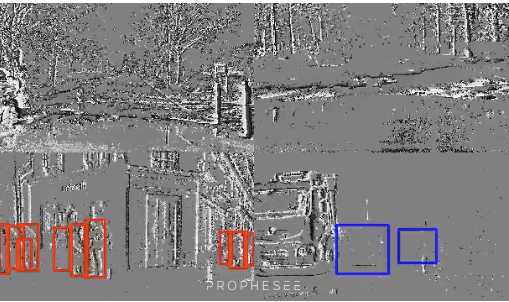}
    \caption{Visualizations of DVS images from CARLA and GEN1.}
    \label{fig:dvs0}
\end{figure}

Figure \ref{fig:dvs0} shows the visualization of the DVS data in CARLA and GEN1. They both are transformed from event sequences. Each event can be an ON or OFF event, depending on the brightness change direction in the corresponding pixel location. Using colors to represent these events gives a visualization like Figure \ref{fig:dvs0}. Typically three colors are used in the visualization, which represent ON event, OFF event, and NO event.

Observing these two visualizations, we can get an idea of what these data are suitable for. The relative movement lets the road-side buildings generate a lot of consecutive spikes on corresponding pixels. Even though these buildings are not directly related to decision making in driving, they can be useful for state estimation or localization. The road marking information is important to vehicles, but DVS data can not clearly capture this information, especially when driving straightly. On the other hand, DVSs can clearly show moving objects and appearance-changing objects. And the sensitivity is robust to bad lighting conditions.

\subsection{Task}
Relying only on DVS data for autonomous driving may not be enough to handle diverse scenarios on the road. This paper designs a simple task to test the feasibility of using SNNs on DVS data for driving. The vehicle is making stop-or-drive decisions based on a sequence of traffic-light observations. To simplify the task, the sequence here only includes two consecutive observations. The traffic light position on the image will not be given by some object detector. An SNN model will be built to map the original DVS observation sequence to stop-or-drive decisions. A CNN model that maps RGB observation sequence to stop-or-drive decisions will also be built and compared with the SNN model.
\begin{figure}[ht]
    \centering
    \includegraphics[width=0.9\linewidth]{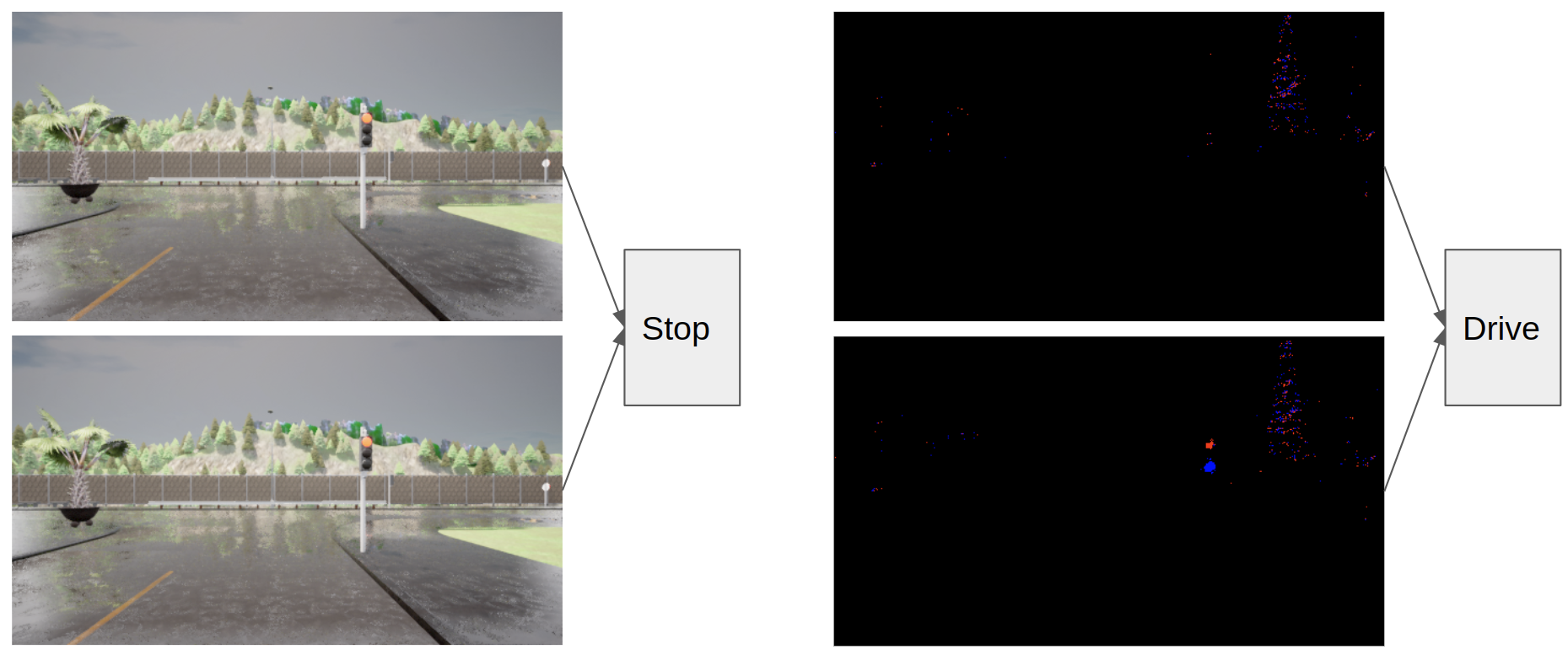}

    \caption{The task investigated in the project}
    \label{fig:dvs}
\end{figure}
\begin{figure*}[ht]
\centering
  \includegraphics[width=0.8\textwidth]{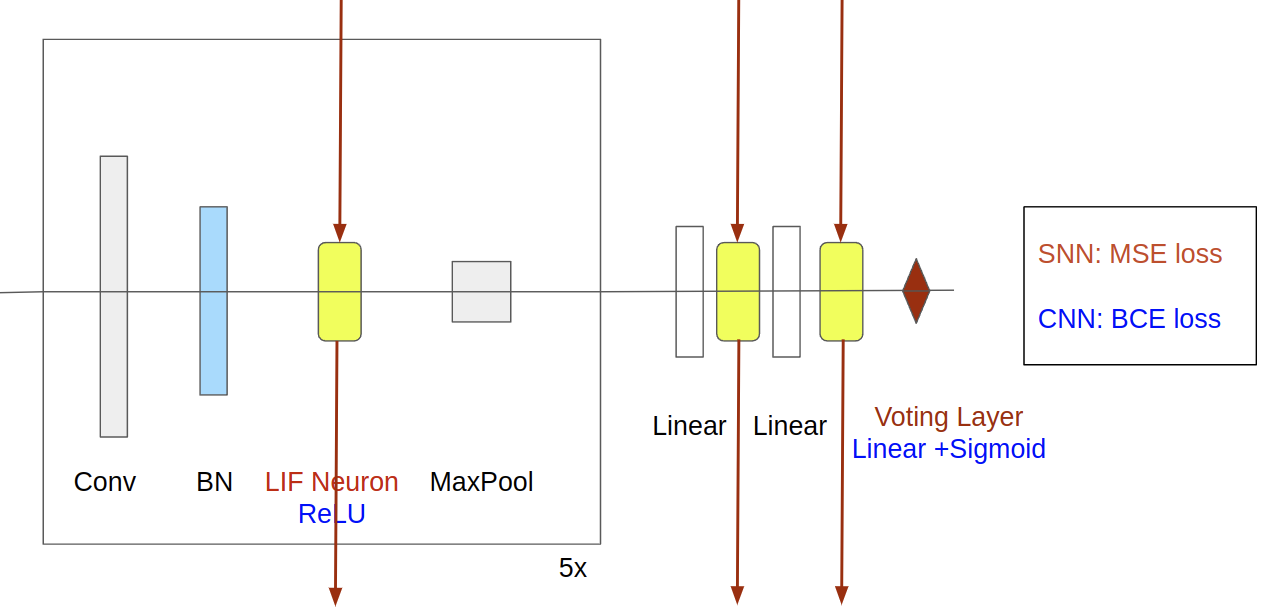}
  \caption{Network structure of the proposed model.}
  \label{fig:stru}
\end{figure*}

\section{METHODOLOGY}
\subsection{Network Structure}
The neural network structure of the proposed method here is based on the neural network designed for the DVS128 Gesture dataset published in \cite{fang2021incorporating}. In this network structure, five repeated blocks composed of a Conv layer, a BatchNormalization layer, a ReLU or LIF Neuron layer, and a MaxPooling layer can downsample the input. Two linear layers and activation neurons can map the flattened tensor to the binary classification result. The CNN version is standard, consisting of commonly used components. Some special components in the SNN version are introduced as follows:

\textbf{LIF Neuron:} Time is a continuous variable, but the neural network processes information received at discrete time points. So we derive a discrete version of the ordinary differentiable equation in Equation (1) as follows
\begin{equation}
    H_t = V_{t-1} + \frac{1}{\tau}(E_r-V_{t-1}+X_t),
\end{equation}
where $I/g_L$ is replaced by the input variable $X_t$ to simplify the notation. $E_r$ denotes the rest-state potential of the neuron. $H_t$ denotes the potential of the neuron before any possible triggering of spikes. $V_t$ denotes the potential after any possible triggering of spikes. $H_t$ and $V_t$ have the following relationship:
\begin{equation}
    V_t = H_t(1-S_t) + E_rS_t,
\end{equation}
where $S_t$ denotes whether a spike is triggered or not. If we set $E_r$ to be 0, we can transform Equation (4) to
\begin{equation}
    H_t = (1-\frac{1}{\tau})V_{t-1} + \frac{1}{\tau}X_t.
\end{equation}
This shows LIF Neuron use a parameter $\tau$ to control the temporal information flow. This also reminds us of the gating mechanism in RNN. This effect is depicted in Figure \ref{fig:stru} using red arrows. Different neurons have different constants $\tau$, which suggests that $\tau$ as a parameter of the LIF Neuron be trainable.

\textbf{MaxPool}: The MaxPooling layer in SNNs is the same as that in other CNNs. However, the MaxPooling layer after the LIF Neuron mimics the winner-take-all mechanism, which allows the fired neuron to communicate with its following layers without considering its neighborhood.

\textbf{Voting layer:} The voting layer after the final LIF Neuron is used to get a final binary classification. This voting layer is essentially an average pooling among all final LIF Neuron outputs. The average pooling operation also improves the model's robustness.

\subsection{Training objective}
The CNN version uses classical binary cross-entropy (BCE) loss, which comes from logistic regression. The training objective for the SNN is the mean squared error (MSE) loss because empirical results show better results with MSE loss than with BCE loss.

\section{EXPERIMENTS}
\subsection{Experiment Design}
Three experiments will be conducted using different combinations of sensors and networks. As the main objective of this project, an SNN with DVS data will be trained and tested. For comparison and further evaluation, a CNN with DVS data and a CNN with RGB data are also trained and tested.

Robustness and generalizability are important properties that we desire a learned model to have. Therefore, both in-domain and out-of-domain testing data will be used to evaluate the learned model.

In the experiments, we want to answer the following several questions:
\begin{itemize}
    \item \textbf{Q1:} Are SNNs more favorable than CNNs on tasks using DVS data?
    \item \textbf{Q2:} Are DVS data better than RGB data on the proposed traffic light task?
    \item \textbf{Q3:} Are SNNs more computationally efficient than CNNs?
\end{itemize}
\subsection{Data Collection}
DVS data and RGB data are collected from the CARLA simulator. The training data cover four different weather conditions. For each weather condition, five segments with four frames per segment are collected. Each segment can generate three double-frame sequences. In total, training data contains sixty double-frame sequences. An example of a double-frame sequence is visualized in Figure 2. For the same four weather conditions, in-domain testing data contain twelve double-frame sequences. Collected on a new weather condition, out-of-domain testing data contain eighteen double-frame sequences.

All data are collected from Town01 in CARLA. It seems that the data are very scarce. But Town01 is a small map, and these data have already covered all the traffic lights in it. The variation in the data comes from other agents and the weather conditions.

\subsection{Implementation Details}
We implemented the proposed SNN model\footnote{The data and code can be found at: \url{github.com/xueleichen/snn-dvs-carla}} and its CNN counterpart using PyTorch and SpikingJelly. As for hyperparameters, the learning rate is 0.001, and the batch size is 12.

Since the CARLA simulator follows the right-hand traffic rule, traffic lights will always appear on the right half-space of the image. The data loader will feed only the right half tensor to the neural network during training and testing.

The surrogate gradient method is used at LIF Neuron nodes. During the back-propagation, the gradient of $arctan()$ is used as the surrogate gradient, which can be formulated as
\begin{equation}
    g'(x) = \frac{\alpha}{2(1 + (\frac{\pi}{2}\alpha x)^2)}
\end{equation}
\subsection{Results and Analysis}
\begin{figure}[ht]
\centering
  \includegraphics[width=0.49\textwidth]{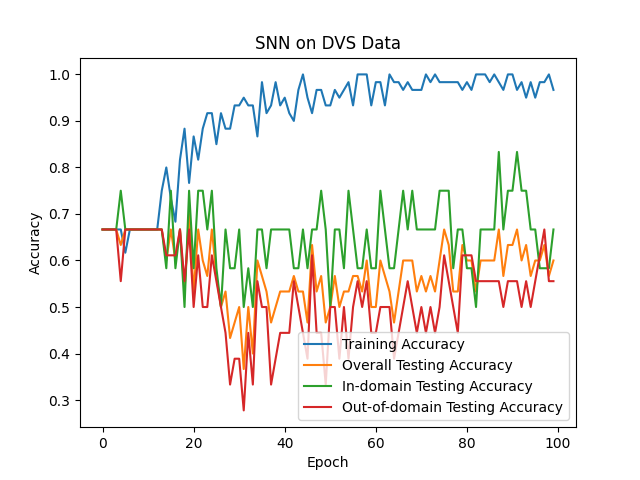}

  \caption{Training and testing accuracy curves of SNN with DVS.}
  \label{fig:snn-dvs}
\end{figure}
\begin{figure}[ht]
\centering

  \includegraphics[width=0.49\textwidth]{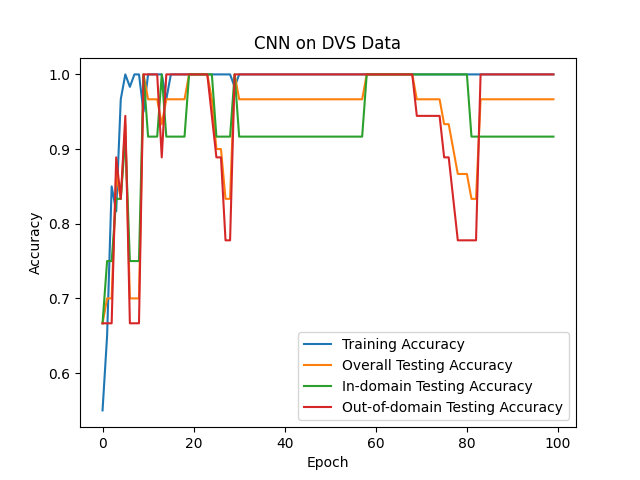}

  \caption{Training and testing accuracy curves of CNN with DVS.}
  \label{fig:cnn-dvs}
\end{figure}
\begin{figure}[ht]
\centering
  \includegraphics[width=0.49\textwidth]{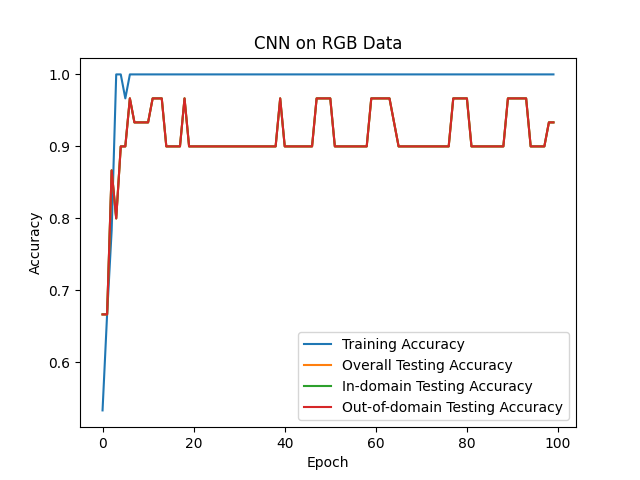}
  \caption{Training and testing accuracy curves of CNN with RGB.}
  \label{fig:cnn-rgb}
\end{figure}
Figures 4-6 show the training curves obtained from different models using different sensor data. All the blue curves in the three plots converge gradually to 1, which shows the gradient-based optimization works on both the SNN and the CNN. The network can learn some features in the data to make correct classifications, at least on training data. The orange line is the curve for overall accuracy considering both in-domain and out-of-domain testing data. The green line is the curve for in-domain testing accuracy. The red line is the curve for out-of-domain testing accuracy. In the following, We will connect the results to the research questions asked in the experiment design section.

Comparing Figure 4 and Figure 5, we can see that the SNN can not achieve acceptable testing accuracy no matter whether the data is out-of-domain or not. However, CNN on DVS data shows good results, with all three testing accuracies larger than 0.9 after some epochs. For \textbf{Q1}, we did not find evidence of SNN being better than CNN, at least on our traffic light task. And actually, on our task, the CNN outperforms the SNN by a large margin.

DVSs are known for detecting changes. Our task is essentially a change detection task that aims to detect the traffic light turning green from red effectively. Comparing the orange curves in Figure 5 and Figure 6, the orange curve in Figure 5 maintains a high value with fewer oscillations, which proves the advantage of DVS data on this task. Comparing the green and red curves in Figure 5 and Figure 6, we can see the CNN with DVS data can reach close to 1, while the CNN with RGB data can never reach that high value. The answer for \textbf{Q2} is that DVS data are more suitable for this traffic light task than RGB data. 

Table 1 shows the computational time of different methods. We can see that the CNN method is faster than the SNN method. This is caused by the special inference mechanism of the SNN. SNN runs like RNN, and every single timestep data needs to be fed into the network to get the final output. However, CNN methods just use stacked input. In addition, LIF Neuron in SNNs is also more time-consuming than ReLU in CNNs. On the other hand, the CNN with DVS is faster than the CNN with RGB. The reason is that the DVS data are very sparse and have binary polarity values. These help reduce the calculation complexity. The results here can not give an answer to \textbf{Q3}.
\begin{table}[]
    \centering
    \begin{tabular}{|c|c|c|c|}
    \hline
         & SNN w/ DVS & CNN w/ DVS & CNN w/ RGB \\
         \hline
         Inference Time(s)& 0.0059 & 0.0015 & 0.0021\\
         \hline
    \end{tabular}
    \caption{Computational time of one inference using different methods}
    \label{tab:time}
\end{table}

It is worth noting that the training-testing accuracy curves shown here are not a thorough evaluation of the three configurations used in experiments. More metrics like F1 score, AUC, and confusion metrics should be considered in future work.

\section{CONCLUSIONS}
This paper proposed to use SNNs on DVS data to solve autonomous driving tasks. During the investigation of DSV data, We observed that DVS data are not enough to independently guide driving behavior because road marking information from DSV data is extremely noisy. A simplified traffic light change detection task is then proposed to evaluate SNN/CNN methods on DVS/RGB data. The results are not as expected. SNNs with DVS got the worst result. Other results only show DVS data are better than RGB data on change detection tasks. We believe that the SNN method can excel under specific conditions or when combined with hardware acceleration. Another possible reason is that this paper uses a hybrid neural network with both convolutional layers and spiking neurons, which may not be the perfect choice. Other SNNs need to be tested in the future.

Not only the traffic light task, but all the tasks in autonomous driving are safety-critical tasks. One modality of the sensor data is never enough to solve autonomous driving tasks. Another promising future direction is to fuse DSV data with many other different sensors to achieve more accurate perception and decision-making.

\addtolength{\textheight}{-12cm}   









 \bibliographystyle{IEEEtran}
 \bibliography{ref}

\end{document}